\documentclass[conference]{IEEEtran}
\IEEEoverridecommandlockouts
\usepackage{cite}
\usepackage{import}
\usepackage{amsmath,amssymb,amsfonts}
\usepackage{algorithmic}
\usepackage{graphicx}
\usepackage{textcomp}
\usepackage{xcolor}
\usepackage{hyperref}
\usepackage{gensymb}
\usepackage[font=footnotesize]{caption}
\usepackage{float}

\def\BibTeX{{\rm B\kern-.05em{\sc i\kern-.025em b}\kern-.08em
    T\kern-.1667em\lower.7ex\hbox{E}\kern-.125emX}}
\begin{document}

\title{Autoregressive Surrogate Modeling of the Solar Wind with Spherical Fourier Neural Operator}

\author{
\IEEEauthorblockN{Reza Mansouri\IEEEauthorrefmark{1},
Dustin J. Kempton\IEEEauthorrefmark{1},
Pete Riley\IEEEauthorrefmark{2},
and Rafal A. Angryk\IEEEauthorrefmark{1}}
\IEEEauthorblockA{\IEEEauthorrefmark{1}Dept. of Computer Science, Georgia State University, Atlanta, GA, United States}
\IEEEauthorblockA{\IEEEauthorrefmark{2}Predictive Science Inc., San Diego, CA, United States}
\IEEEauthorblockA{Email: rmansouri1@student.gsu.edu}
}

\maketitle

\begin{abstract}

The solar wind, a continuous outflow of charged particles from the Sun’s corona, shapes the heliosphere and impacts space systems near Earth. Accurate prediction of features such as high-speed streams and coronal mass ejections is critical for space weather forecasting, but traditional three-dimensional magnetohydrodynamic (MHD) models are computationally expensive, limiting rapid exploration of boundary condition uncertainties. We introduce the first autoregressive machine learning surrogate for steady-state solar wind radial velocity using the Spherical Fourier Neural Operator (SFNO). By predicting a limited radial range and iteratively propagating the solution outward, the model improves accuracy in distant regions compared to a single-step approach. Compared with the numerical HUX surrogate, SFNO demonstrates superior or comparable performance while providing a flexible, trainable, and data-driven alternative, establishing a novel methodology for high-fidelity solar wind modeling. The source code and additional visual results are available at \url{https://github.com/rezmansouri/solarwind-sfno-velocity-autoregressive}.

\end{abstract}


\section{Introduction}

\begin{figure}[!t]
    \centering
    \includegraphics[width=1\linewidth]{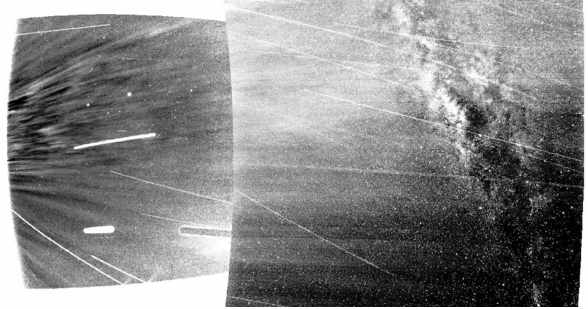}
    \caption{
    Parker Solar Probe WISPR view at $\approx18R_\odot$, August 2021. The image captures the solar wind and coronal streamers (left) from within the Sun's atmosphere, with the Milky Way visible on the right, looking away from the Sun. Streaks are due to energetic particle impacts (Credit: NASA/Johns Hopkins APL/Naval Research Laboratory).
}

    \label{fig:mas_1}
    
\end{figure}

The solar wind is a constant stream of charged particles, mainly protons and electrons, that flows out from the Sun's outer atmosphere, known as the corona. It is pushed outward by heat and magnetic pressure, moving along open magnetic field lines, especially those from coronal holes (see fig.~\ref{fig:mas_1}). As it travels at speeds of more than a million miles per hour, it carries the Sun's magnetic field with it and creates the heliosphere, a large region filled with plasma that surrounds the solar system \cite{nasasolarwind}. The solar wind changes over time, with both fast and slow streams, as well as sudden events like coronal mass ejections (CMEs), which can disturb the space environment near Earth. These disturbances can cause geomagnetic storms that affect satellites, GPS systems, and power grids. Predicting the solar wind is especially important during extreme events such as the Carrington event of 1859 \cite{nrc2009, carrington1859description}.

There are several numerical models used to simulate how the solar wind travels from the Sun (30 $R_\odot$) to Earth (1 AU). These models vary in how simple, fast, or physically detailed they are. Some examples include the ballistic extrapolation method \cite{nolte1973, neugebauer1998}, the Arge–Pizzo kinematic model \cite{argepizzo}, the one-dimensional upwind scheme \cite{riley2011}, the HUX (Heliospheric Upwinding eXtrapolation) method \cite{hux1, hux2}, and full three-dimensional magnetohydrodynamic (MHD) models. Each model is suited for different situations, depending on the available computing resources and the level of accuracy needed. Among these, the MAS (Magnetohydrodynamics Algorithm outside a Sphere) model~\cite{mas} is the most physically detailed and well-tested. It solves the full three-dimensional resistive MHD equations and has demonstrated the ability to reproduce essential characteristics of solar wind flow and stream interactions.

Machine learning (ML) based models are becoming important tools for modeling different physical systems. In our case, surrogate models are valuable for enabling ensemble forecasting at reduced computational cost, making it practical to explore a wide range of conditions and uncertainties. Standard MHD solvers work by dividing space into a grid and solving complex partial differential equations (PDEs), which makes high-resolution simulations of the solar wind and CMEs time-consuming and expensive. ML surrogates aim to capture the key behavior of the system well enough for forecasting and estimating uncertainty, without needing to solve the full physical equations every time \cite{gramacy2020surrogates}. This is especially useful in high-dimensional problems, such as predicting the time and strength of a CME's arrival, where running many traditional simulations would be computationally prohibitive \cite{mays2015:ensemble, riley2012:geoeffectiveness}. Since running large ensembles of high-resolution MHD simulations is computationally expensive, ML surrogates offer a more practical and scalable way to generate predictions at a reduced cost.


In this study, we investigate a machine learning surrogate based on the Spherical Fourier Neural Operator (SFNO)~\cite{bonev:sfno} with an autoregressive functionality. The model is trained to predict the radial velocity field \(v_r\), using simulation data from the MAS MHD model~\cite{mas} as the reference. Section~\ref{sec:background} provides an overview of related research on solar wind modeling and ML surrogates used in physical systems. Section~\ref{sec:methods} describes our approach, including the dataset, modeling steps, the autoregressive training procedure, and evaluation metrics. Section~\ref{sec:results} presents the results and discusses how well the model performs, how well it generalizes, and how it can be useful for real-world space weather forecasting.

\section{Background}
\label{sec:background}

Recent studies in data-driven solar wind forecasting have used deep learning models that rely on either time-based data or images to improve prediction accuracy. WindNet~\cite{upendran:windnet} forecasts the daily average solar wind speed by analyzing EUV images of the solar corona, particularly the 193~\AA{} and 211~\AA{} wavelengths, and identifies important features such as bright areas near coronal holes. Another model, the Solar Wind Attention Network (SWAN)~\cite{cobos:swan}, uses a fully image-free, autoregressive method with an attention-based encoder–decoder design. SWAN achieves a lower RMSE compared to WindNet~\cite{upendran:windnet} and earlier convolutional neural network (CNN) based models. Moreover, mapping granular structures through semantic segmentation, for example, presents a potential avenue for linking small-scale solar dynamics to surrogate models of the solar wind~\cite{imax}.

Unlike models that predict average values over space, our goal is to capture the complete spatial development of the solar wind throughout the heliosphere. To do this, we need to map physical states while preserving the symmetries and fundamental physics of the system. Conventional deep learning models, tied to fixed input–output grids, are not well suited for this. Thus, we use an operator learning approach in the spectral domain that better reflects how the solar wind propagates.

Operator learning~\cite{operator, deeponet} has become an effective approach for modeling relationships between infinite-dimensional function spaces, making it well suited for problems governed by physical laws. The Fourier Neural Operator (FNO)~\cite{li:fno} learns these mappings by parameterizing the integral kernel in Fourier space, allowing for both flexibility and efficiency. It has achieved top accuracy on PDE benchmarks such as Burgers’ equation and the Navier–Stokes equations. Additionally, it is the first learning-based method capable of modeling turbulent flows with zero-shot super-resolution.

Neural operator models have recently shown strong potential for efficiently solving complex physical systems. GL-FNO~\cite{glfno} accelerates coronal magnetic field simulations by over $\times20,000$ while capturing both global and fine-scale features. PINO~\cite{pino} enables real-time extrapolation of 3D nonlinear force-free fields from 2D magnetograms with comparable accuracy to conventional methods. Fourier-MIONet~\cite{fouriermionet} predicts long-term multiphase flow in porous media with high accuracy and low computational cost.  FourCastNet~\cite{pathak:fourcastnet} is a data-driven Fourier neural network for accurate short- and medium-term weather forecasts, especially for high-resolution variables, with fast predictions enabling large ensembles. These works highlight the versatility of neural operators across astrophysical and geophysical applications.

Spherical Fourier Neural Operator (SFNO)~\cite{bonev:sfno} adapts the FNO framework to spherical domains by replacing the Fast Fourier Transform (FFT) with the Spherical Harmonic Transform (SHT). This change ensures the model respects rotational symmetry and improves its stability. As a result, SFNOs are well suited for learning spatial or spatio-temporal operators on spherical surfaces. Section~\ref{sec:methods} explains the modeling approach and experimental details.

\section{Methods}
\label{sec:methods}

\subsection{Dataset}
Our dataset contains 616 Carrington rotations between 19 February 1975 and 1 January 2025, covering more than four complete solar cycles. The data come from MAS simulations that use boundary conditions at $30\ R_\odot$, derived from observations made by three instruments: the Kitt Peak Observatory (KPO), the {SOHO}/Michelson Doppler Imager (MDI), and the {SDO}/Helioseismic and Magnetic Imager (HMI).

\begin{figure}[!h]
    \centering    \includegraphics[width=1\linewidth]{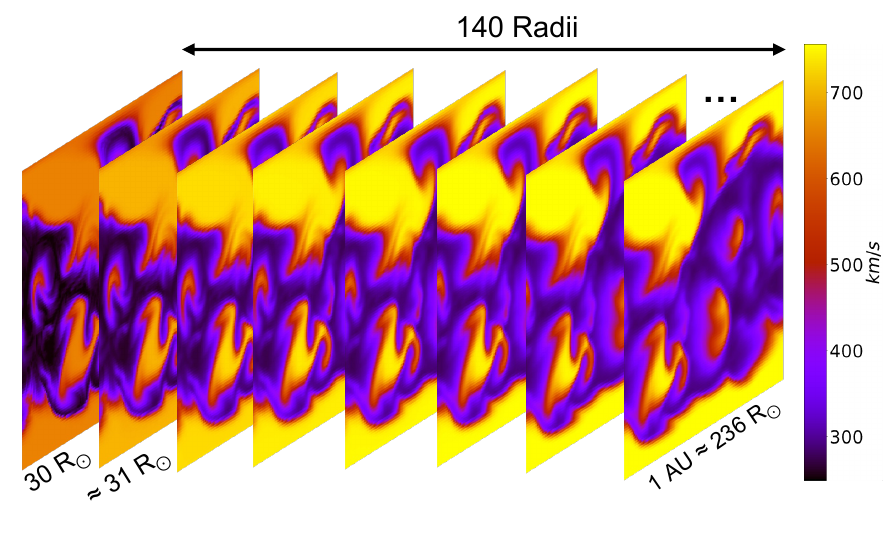}

\caption{
Equiangular projections of radii for one instance (MAS simulation of CR 2285, 2–29 June 2024). The first radius \(r_0 = 30\,R_\odot\) is the input; the remaining 139 radii are the ground truth. The physical area grows rapidly with radius, so scale increases substantially at larger distances.
}
    \label{fig:mas_2}
\end{figure}

The dataset contains a total of {1,069} samples from different instruments and Carrington rotations. Each sample is a three-dimensional cube on a medium-resolution grid of $(140, 111, 128)$, representing (radius, latitude, and longitude). Along the radial axis, the sampling step is approximately $1.48\ R_\odot$, providing uniform coverage from $30\ R_\odot$ to 1~AU. The first radial slice ($r_0$) of each cube serves as the input, describing the velocity field at the inner boundary of the heliosphere simulation. The remaining 139 radial layers ($r_1$ to $r_{139}$) act as the ground truth, representing the solar wind velocity structure farther into the heliosphere, making this an extrapolation problem. (see fig.~\ref{fig:mas_2}).

\subsection{Modeling Pipeline}

We use the SFNO\cite{bonev:sfno}, which uses Spherical Harmonic Transform (SHT), preserving symmetries and rotational equivariance. The model architecture includes an encoder Multilayer Perceptron (MLP), multiple SFNO blocks that combine pointwise MLPs with spectral convolutions, and a decoder MLP. Skip connections are used to support stable autoregressive rollout.

In our setup, we use 110 modes in latitude, the maximum allowed by polynomial quadrature rules based on Gauss–Legendre sampling~\cite{gauss}, which corresponds to 111 Gauss–Legendre grid points. For the longitudinal direction, periodic over $[0, 2\pi]$, we use 64 modes, the maximum permitted by the Shannon–Nyquist sampling theorem~\cite{shannon}, corresponding to 128 equispaced longitudinal points. The model takes one input channel representing the solar wind radial velocity at the inner boundary (\(r_0 = 30\,R_\odot\)) and produces 139 output channels covering the range from \(r_1 \approx 31\,R_\odot\) to 1~AU. No tensor factorization is applied in the SFNO layers to keep the full learning capacity. The number of SFNO layers and the number of hidden channels in each layer are treated as tunable hyperparameters.

\subsection{Baseline Numeric Model (HUX-f)}  
As a baseline solver for our study, we use the Heliospheric Upwind eXtrapolation (HUX) model~\cite{hux1, hux2}, which is a simplified physics model for the solar wind’s radial velocity \(v_r\) beyond 30 solar radii (\(R_\odot\)). HUX reduces the fluid momentum equation by assuming radial flow and ignoring pressure and gravity effects, allowing efficient propagation with upwind finite-difference methods. In practice, HUX-f advances the solution radially by iteratively updating each step from the previous one, making it a forward-marching numerical solver rather than a predictive model. The model can run forward (HUX-f), from \(30\,R_\odot\) to 1~AU, or backward (HUX-b), tracing the solar wind radial velocity from 1~AU back to \(30\,R_\odot\). In this study, we use HUX-f as our numerical baseline and compare its results with those of our data-driven SFNO model.

\subsection{Training Strategy}

\subsubsection{Loss Function and Optimization}
The model is trained using a layer-wise two-dimensional \(L_{2}\) loss, defined as
\begin{align}
\mathcal{L}_{2}^{(2\mathrm{D})}
&= \frac{1}{BC} \sum^{B\times C}_{b,c}
\left(
\sum^{H\times W}_{i,j}
\lvert y_{bcij} - \hat{y}_{bcij} \rvert^2
\right)^{1/2}
\label{eq:l2_2d_short}
\end{align}
where \(B\) is the batch size, \(C\) is the number of output channels (number of radii), \(H \times W\) are the latitude and longitude grid dimensions, and $y$ and $\hat{y}$ are the ground truth and the prediction respectively. We use the Adam optimizer with a fixed learning rate of \(8\times10^{-4}\), and train with a batch size of 32. The model checkpoint that gives the lowest validation loss is selected as the final model.

\subsubsection{Cross-Validation and Training}
To select the best architecture, we perform 150-epoch 5-fold cross-validation using MSE as the criterion, using 4- and 8-layer SFNOs with 64, 128, and 256 channels. The best model is retrained on the full training set for 200 epochs and evaluated on the hold-out test set. In every experiment, the training set is min-max normalized to $[0,1]$, with predictions rescaled using the training split’s bounds. Table~\ref{tab:dataset} summarizes the dataset split.

\begin{table}[h!]
\caption{Carrington rotation ranges, corresponding date spans, and instance counts used for training, cross-validation, and testing.}
\centering
\resizebox{\columnwidth}{!}{%
\begin{tabular}{l|l|l|l}
\textbf{Phase} & \textbf{CR \#} & \textbf{Date Range} & \# \textbf{Instances} \\
\hline
Training \& CV & 1625--2169 & 19 Feb. 1975 -- 4 Oct. 2015 & 697 \\
Testing & 2170--2293 & 31 Oct. 2015 -- 1 Jan. 2025 & 372 \\
\end{tabular}%
}
\label{tab:dataset}
\end{table}

\subsubsection{Autoregressive prediction}

Our results show that while the model predicts well for radii close to the input boundary, its accuracy and sharpness get worse for larger radii farther out. To improve long-range predictions, we use an autoregressive inference method additional to the current autoregressive functionality of the SFNO. The model predicts a limited number of radii (the predictive horizon, for example 20), then feeds the last predicted radius back as input to forecast the next set. This process repeats until the full radial range is covered.

\begin{figure}[!h]
    \centering
    \includegraphics[width=1\linewidth]{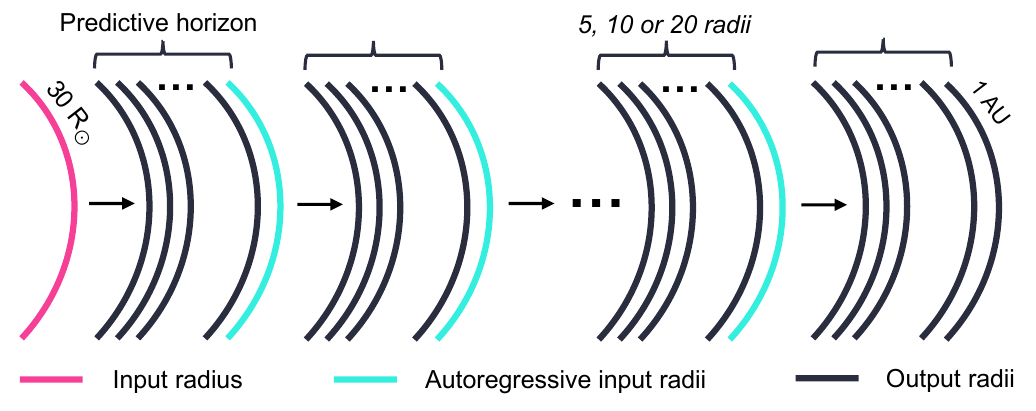}
    \caption{Autoregressive prediction approach. The model predicts a fixed number of radii (predictive horizon) per step, feeding the last predicted radius back as input for the next prediction. This repeats until the full radial range is covered. During training, teacher forcing uses ground truth data as input to stabilize learning.}
    \label{fig:autoregressive}
\end{figure}

During training, we use a teacher-forcing approach~\cite{teacher1, teacher2} in which ground truth MAS data are fed instead of the SFNO’s own predictions for later steps, enabling faster convergence. We train models with the optimal architecture using different predictive horizons—20, 10, or 5 radii—to find the best balance between horizon length and accuracy; these values were chosen empirically to trade off predictive range with training and inference efficiency. Since we keep the total number of predicted slices fixed at 140, one more than the 139 available in the ground truth, the final step produces an extra radius that is discarded. Fig.~\ref{fig:autoregressive} illustrates this method.

\subsubsection{System Specifications}
All models are implemented in Python 3.10.12 with PyTorch version 2.2.1 and the NeuralOperator package~\cite{neuralop1, neuralop2} version 1.0.2. Experiments are run on a system with Ubuntu 22.04.3, an NVIDIA A40 GPU with 48~GB memory, and about 500~GB of RAM.

\subsection{Evaluation metrics}

\begin{figure}[!h]
    \centering
    \includegraphics[width=1\linewidth]{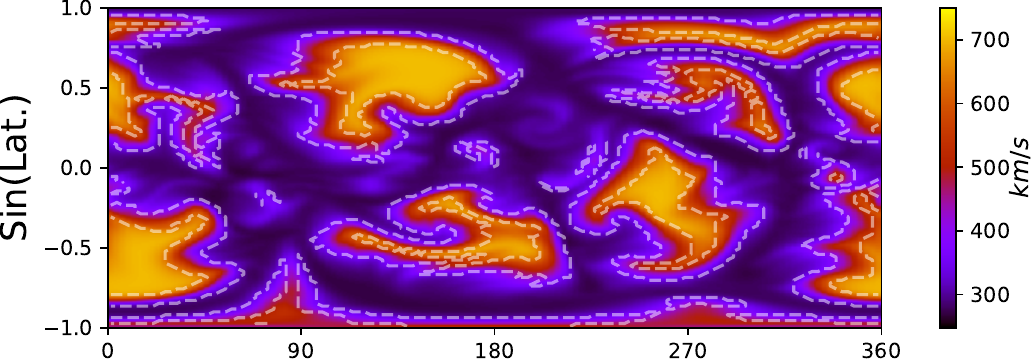}
    \caption{
    Solar wind radial velocity at $\approx49\ R\odot$ for Carrington Rotation 2285 from the MHD solution. The edge regions, detected using a Sobel filter, are outlined with dashed lines.
}
    \label{fig:edges}
\end{figure}

We assess model performance using several complementary metrics. Mean Squared Error (MSE), a standard regression metric, is used during cross-validation to select the best hyperparameters. To better capture high-gradient regions, where fast solar wind can overtake slower wind and create shocks~\cite{riley:tilts, mas}, we also compute an edge-specific MSE. This version of MSE is calculated only over high-gradient regions identified in radial slices using a $3\times3$ Sobel filter (see fig.~\ref{fig:edges}). For the final evaluation, we also report Earth Mover’s Distance (Wasserstein distance) and the Universal Image Quality Index~\cite{uiqi}, both computed as averages over radial slices.

\section{Results and Discussion}
\label{sec:results}

\begin{figure}[!h]
    \centering
    \includegraphics[width=1\linewidth]{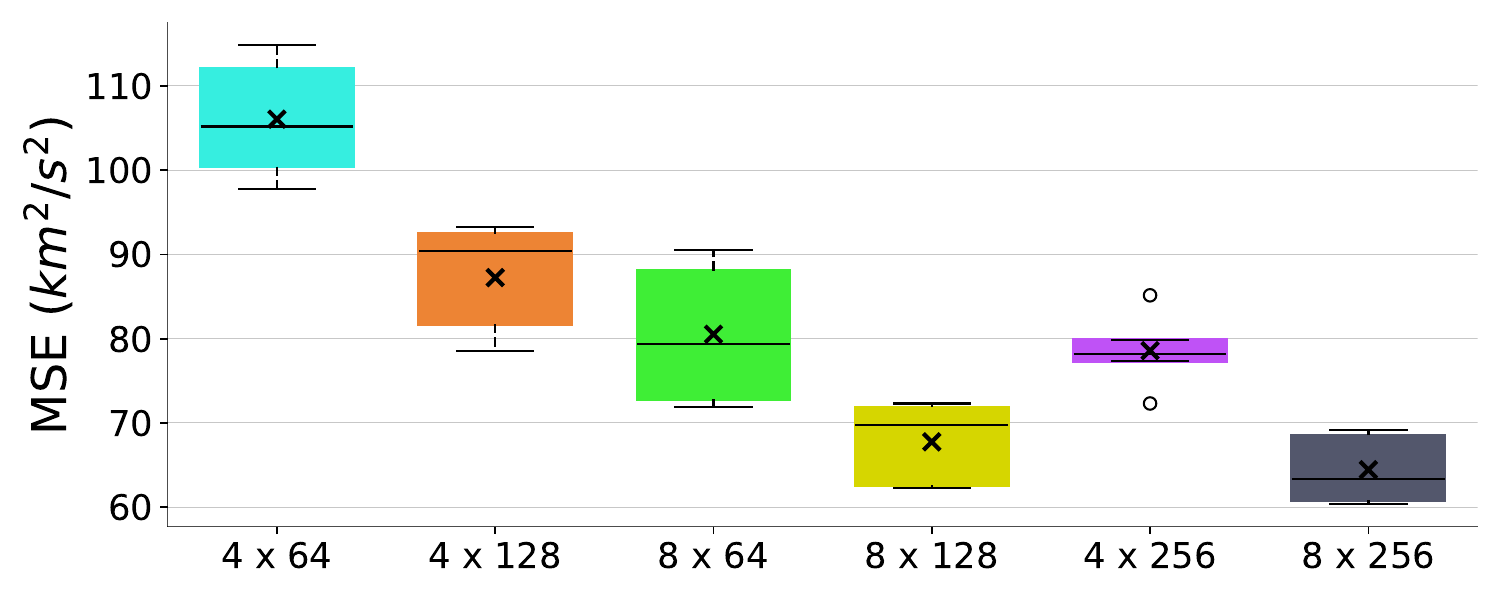}
    \caption{
Cross-validation results showing the MSE across five folds for SFNO architectures with varying depths and hidden channel sizes; for example, $4\times64$ denotes a model with 4 layers and 64 hidden channels.
}
    \label{fig:mse_cv}
\end{figure}

The 8-layer, 256-channel SFNO consistently achieves the lowest MSE across all cross-validation folds, outperforming shallower or narrower variants in capturing the spatial and spectral structure of the solar wind radial velocity field (fig.~\ref{fig:mse_cv}). Based on this robustness and accuracy, we selected it as the final architecture for investigating autoregressive prediction to improve long-range performance.

\begin{figure}[!h]
    \centering
    \includegraphics[width=1\linewidth]{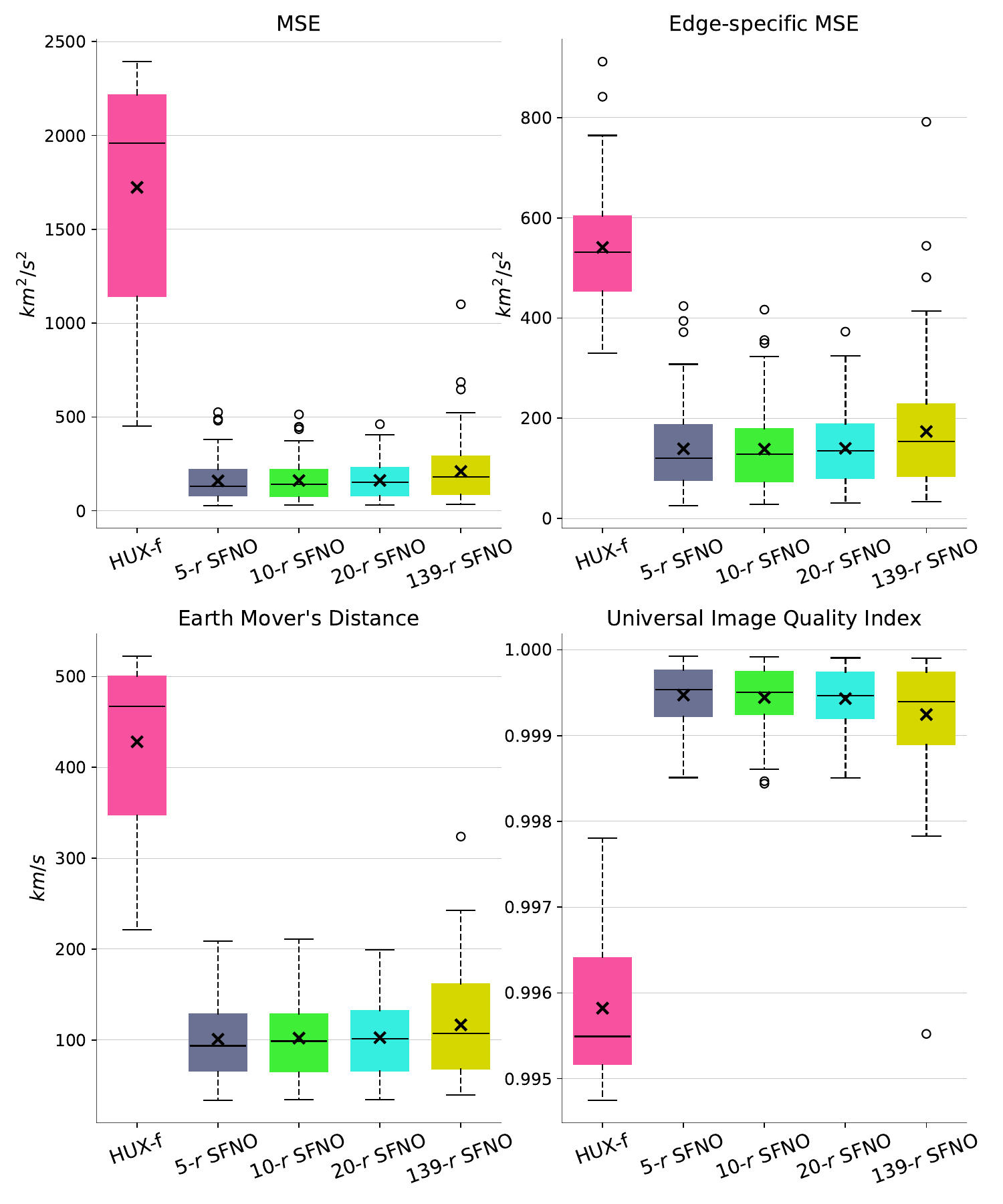}
    \caption{
Performance distributions comparing HUX-f and $8\times256$ SFNO models across key metrics on our test set. On average, SFNO models outperform HUX-f, with the 5-radius SFNO showing best results on average. The small metric differences among SFNOs highlight the need for more suitable evaluation metrics for this physical phenomenon.
}
\label{fig:metrics_boxplot}
\end{figure}

Moving on with the autoregressive approach and the $8\times256$ SFNO, fig.~\ref{fig:metrics_boxplot} shows test set metrics for SFNO models with 5, 10, and 20-radius predictive horizons compared to the HUX-f baseline. On average, SFNO variants outperform HUX-f, with shorter horizons producing more accurate and stable forecasts. The 5-radius model achieves the best overall performance, highlighting the trade-off between prediction horizon and accuracy.

\begin{figure}[!h]
    \centering
    \includegraphics[width=1\linewidth]{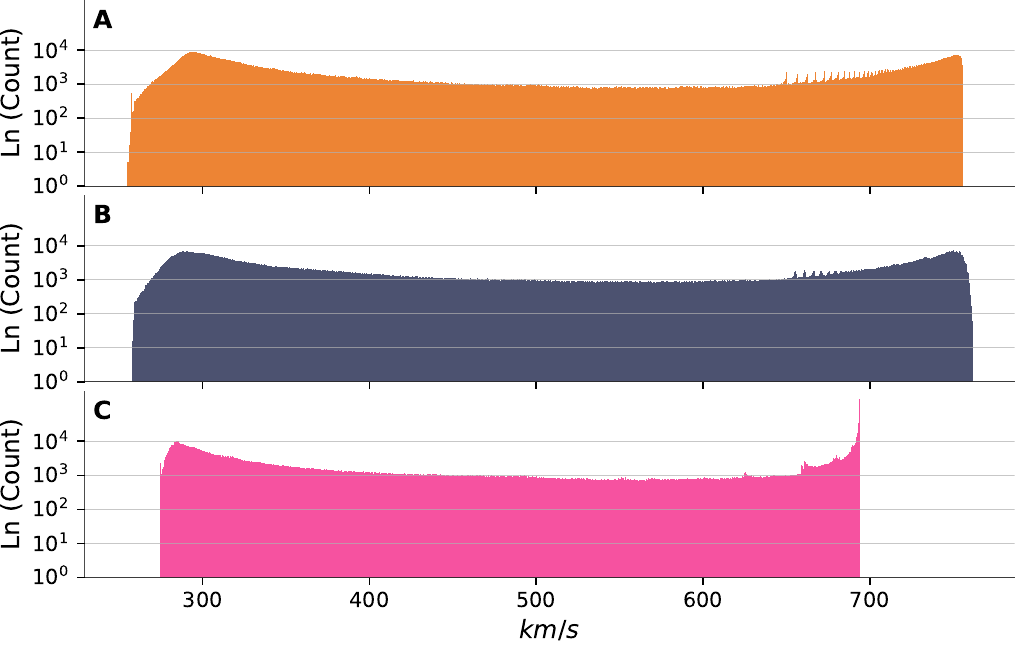}
    \caption{
Distribution of solar wind speed values for Carrington Rotation 2285, shown in log scale. (A) MAS (our ground truth), (B) Our winning 5-radius SFNO, and (C) HUX-f method. The SFNO distribution closely matches the MAS reference, capturing both the high- and low-speed wind regimes more accurately than HUX-f.
}
    \label{fig:hist}
\end{figure}

Fig.~\ref{fig:hist} shows solar wind speed distributions for Carrington Rotation 2285, a period of high solar activity. The SFNO model closely matches the MAS ground truth, capturing both high- and low-speed regimes, while HUX-f deviates and misses some variability. This highlights SFNO’s superior ability over HUX-f to represent solar wind behavior under complex conditions.

\begin{figure*}[!h]
    \centering
    \includegraphics[width=1\linewidth]{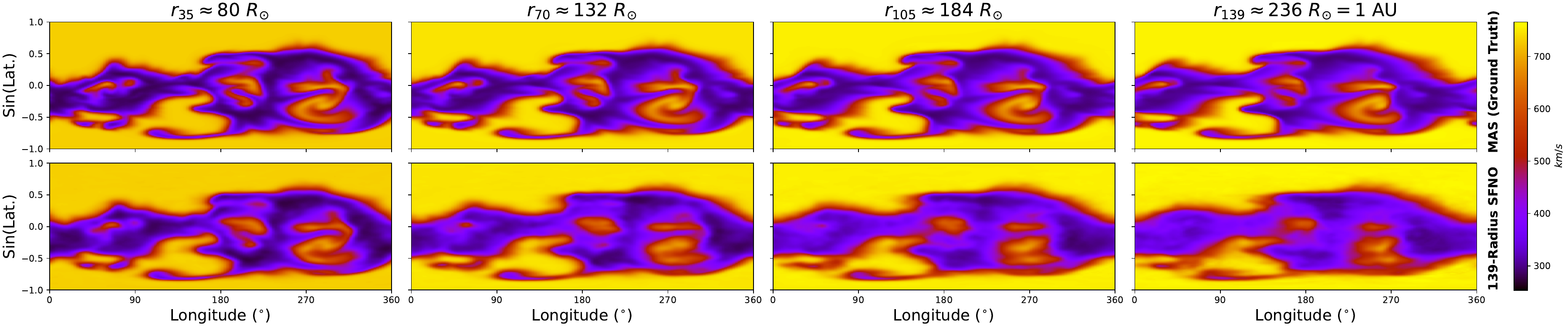}
    \caption{Comparison of the 139-radius (no auto-regression) SFNO prediction (bottom row) with the MAS ground truth (top row) for CR2160 for different radial distances. Results show that while the model predicts well for radii close to the input boundary, its accuracy and sharpness decrease for larger distances farther out; autoregressive inference can improve long-range predictions with smaller predictive horizions.}
\label{fig:poor-139}
\end{figure*}

\begin{figure}[!h]
    \centering
    \includegraphics[width=0.954\linewidth]{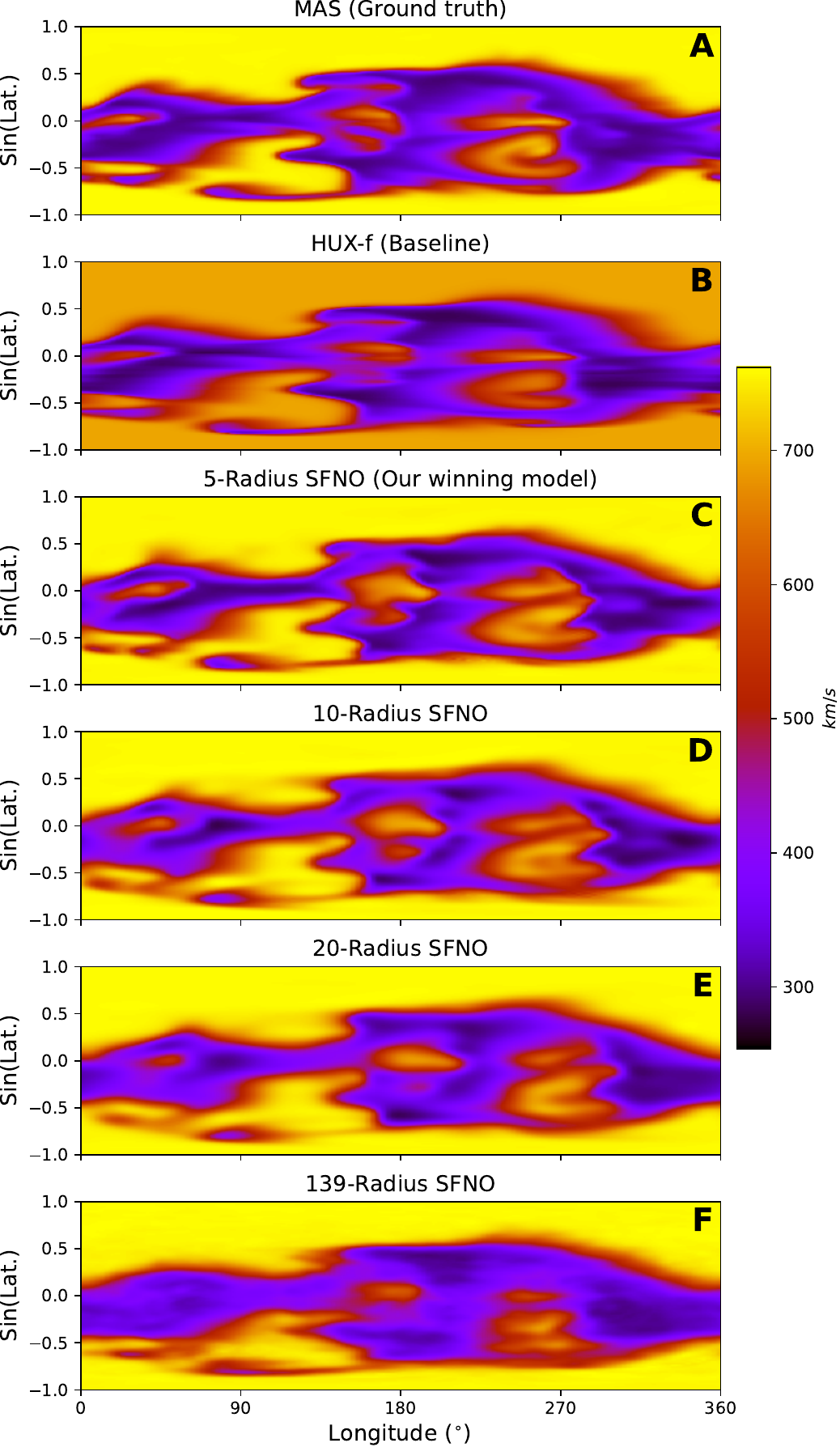}
    \caption{
Solar wind speed at 1~AU for Carrington Rotation 2160. (A) Ground truth from the MHD solution. (B) Estimate from the HUX-f technique. (C), (D), (E) and (F) represent the estimate from the $8\times256$ SFNO model using 5-Radius, 10-Radius, 20 Radii, and 139-Radius predictive horizons. Smaller predictive horizons produce smoother and visually better estimates, though HUX-f remains the smoothest overall. This case represents one of the hardest instances for SFNO based on MSE.
}
\label{fig:visual_comparison}
\end{figure}

\begin{figure}[!h]
    \centering
    \includegraphics[width=0.954\linewidth]{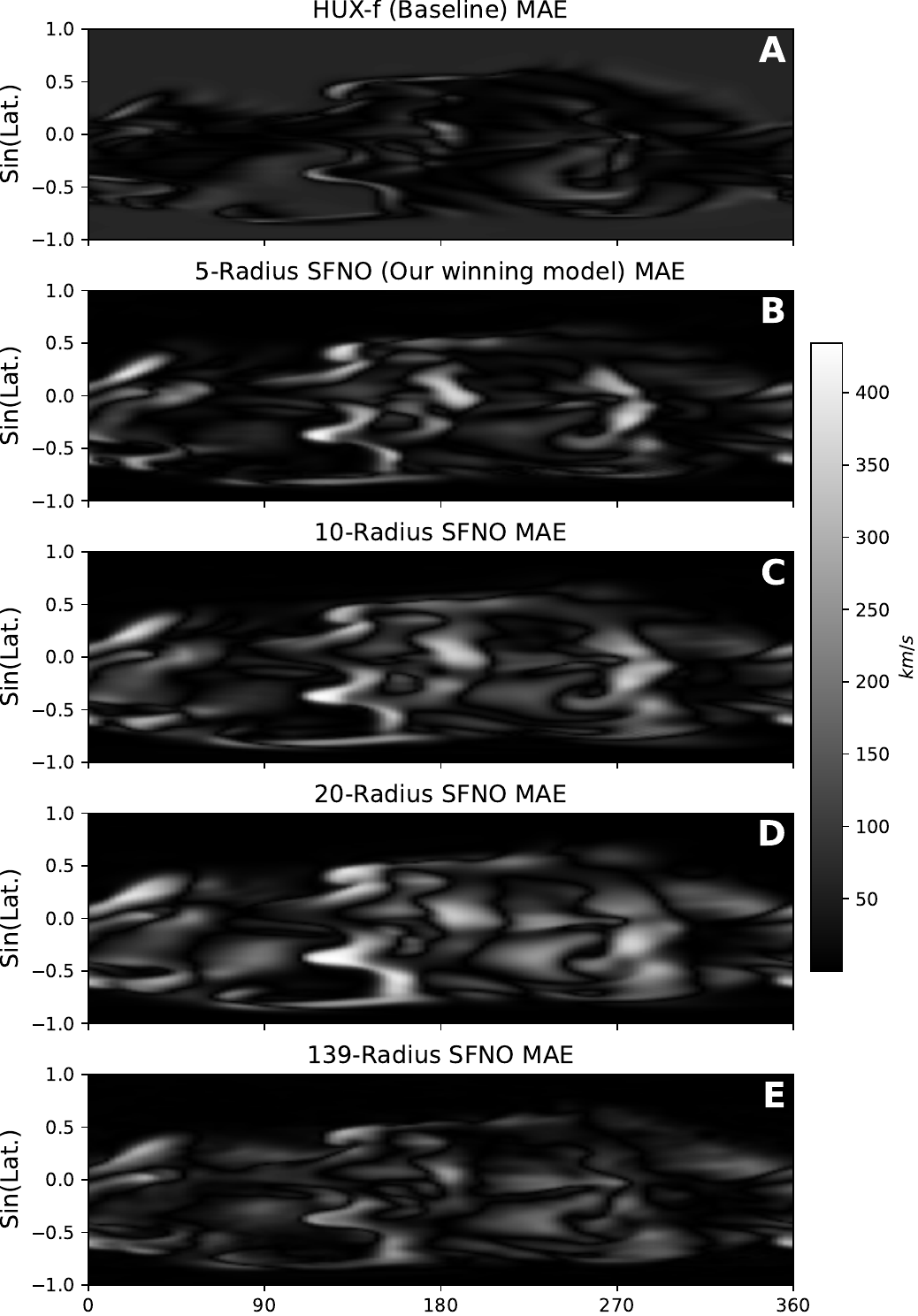}
    \caption{
Mean absolute error at 1~AU for Carrington Rotation 2160 : (A) HUX-f model (baseline), (B) our 5-radius winning SFNO, (C) 10-, (D) 20-, and (E) 139-radius SFNO.
The 5-radius SFNO achieves low global errors but performs worse than HUX-f in high-gradient regions. HUX-f, although sharper in such areas, struggles in polar and high-speed wind regions due to its empirical acceleration term, not derived from first-principles physics.
}
\label{fig:mae}
\end{figure}

SFNO's superiority, particularly over large-scale background fast wind is due to HUX-f using an empirical acceleration term, not derived from first principles, to account for residual solar wind acceleration. Visual comparisons (Fig.~\ref{fig:visual_comparison}) and error maps (Fig.~\ref{fig:mae}) highlight distinctions that standard metrics may overlook.

\begin{figure}[!h]
    \centering
    \includegraphics[width=1\linewidth]{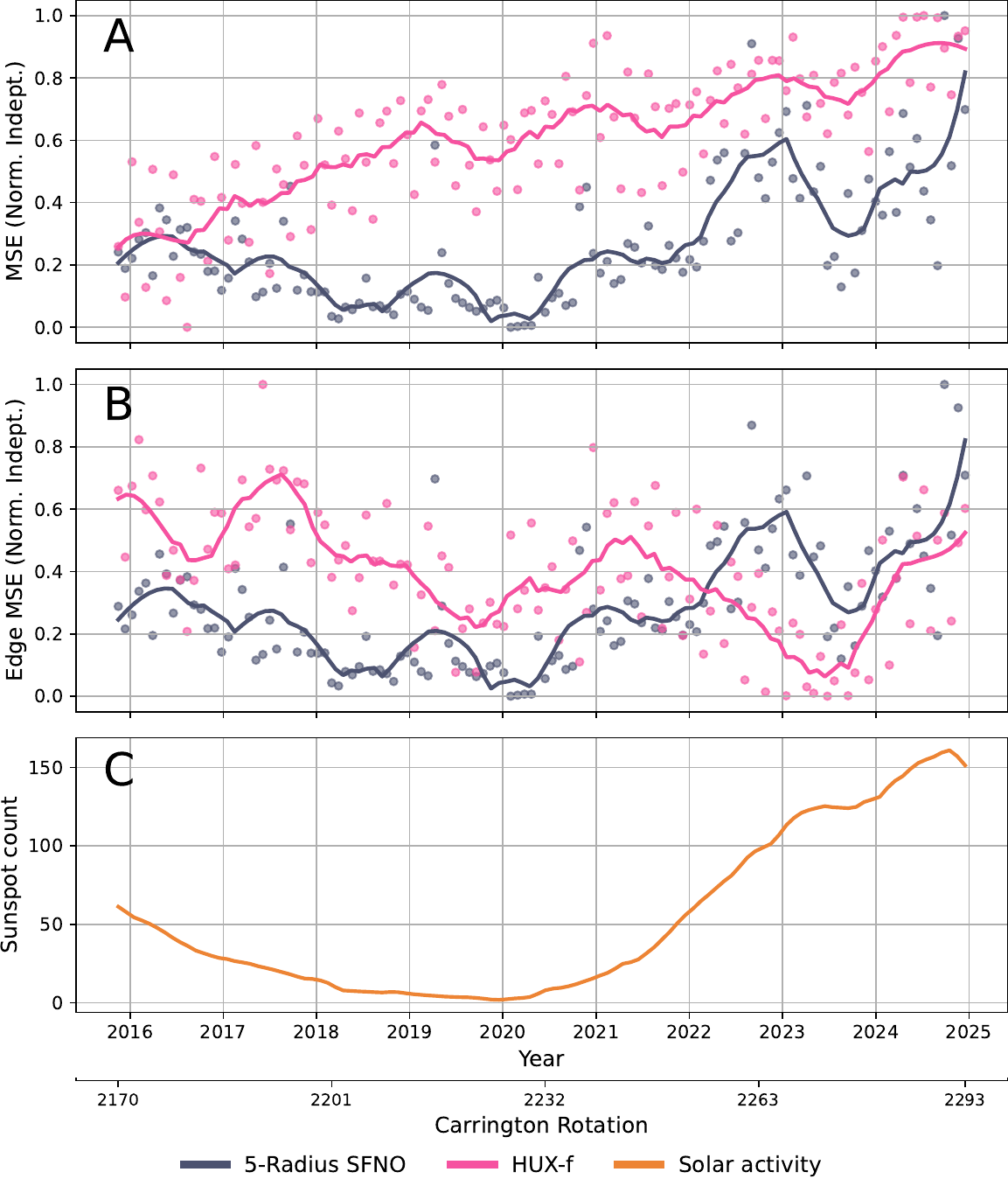}
    \caption{
(A) MSE, (B) edge-specific MSE, and (C) 13-month smoothed sunspot count for the test set. Scatter points show raw model errors per rotation, and solid lines indicate smoothed trends. Errors are min–max normalized independently to highlight temporal patterns; the focus is on curve shapes, not magnitudes. SFNO’s overall and edge-specific MSE closely align, arising mainly in high-gradient regions, and both error trends follow solar activity, showing greater difficulty during active periods.
}
\label{fig:difficulty}
\end{figure}

Finally, we explored the relationship between model prediction errors and solar activity levels, using sunspot counts as an indicator of the solar wind’s complexity. Fig.~\ref{fig:difficulty} compares the mean squared error (MSE) and edge-specific MSE of the 5-radius SFNO and HUX-f models with the 13-month smoothed sunspot count, a standard proxy for solar activity. To focus on temporal variation rather than absolute magnitude, each error curve is independently min--max normalized, mapping its values to the range \([0,1]\). This normalization ensures that the relative shapes of the curves, not their raw magnitudes, can be directly compared, allowing a clearer assessment of how prediction difficulty evolves over time for each model. For SFNO, the close alignment between its MSE and edge-specific MSE indicates that high-gradient regions dominate its errors. Both SFNO and HUX-f show elevated errors during periods of high solar activity, suggesting that the increased structural complexity of the solar wind under active conditions poses a greater challenge to predictive accuracy.

While traditional solvers such as HUX~\cite{hux1, hux2}, which propagate radial velocity one radius at a time while feeding forward their own outputs, our SFNO advances in larger increments of five solar radii. This coarser autoregressive stepping not only succeeds but leads to improved predictive accuracy, confirming the effectiveness of the approach. Overall, these results establish autoregressive SFNO as a strong data-driven alternative to physics-based models, offering a new path forward for solar wind surrogate modeling.

\section*{Conclusion}
The solar wind is a continuous flow of charged particles from the Sun that shapes the heliosphere and impacts space weather near Earth. Accurate modeling of its complex dynamics is crucial for protecting satellites, communications, and power infrastructure. In this work, we developed an autoregressive ML surrogate based on the Spherical Fourier Neural Operator, which improves long-range prediction by iteratively forecasting segments of the solar wind radial velocity. Our model matches or surpasses traditional numerical solvers like HUX, offering a flexible and trainable framework for efficient real-time forecasting. Future work will focus on enhancing evaluation metrics and loss functions tailored to this physical phenomenon, integrating additional variables such as plasma density $\rho$, and exploring physics-informed training to deepen the model’s physical understanding.

\section*{Acknowledgements}

The data used in this work are publicly available at \url{https://predsci.com/data/runs}.  
The code implementation can be accessed at \url{https://github.com/rezmansouri/solarwind-sfno-velocity-autoregressive}. The authors also acknowledge support from NASA’s HSO-Connect program (grant 80NSSC20K1285) and from the PSP WISPR contract NNG11EK11I to NRL, provided through subcontract N00173-19-C-2003 to Predictive Science.


\bibliographystyle{IEEEtran}
\bibliography{refs}

\end{document}